\documentclass[10pt,twocolumn,letterpaper]{article}

\usepackage{cvpr}      % To produce the REVIEW version

\usepackage{soul} % underline
\usepackage{url}
\usepackage[utf8]{inputenc} % standard text encoding
\usepackage{amsmath,amssymb, amsfonts}
\usepackage{booktabs}
\usepackage{algorithm2e}
\usepackage{comment}
\usepackage{lipsum}  
\usepackage{tabularx}
\usepackage{tabulary}
\usepackage{multirow}
\usepackage{csquotes}
\usepackage{graphicx}
\usepackage[dvipsnames]{xcolor}
\newif\ifdraft
\drafttrue

\ifdraft
  \newcommand{\daniel}[1]{{\textcolor{teal}{\textbf{Daniel:}~\enquote{#1}}}}
  \newcommand{\pablo}[1]{{\textcolor{blue}{\textbf{Pablo:}~\enquote{#1}}}}
  \newcommand{\fix}[1]{{\textcolor{red}{\textbf{FIXME:}~\enquote{#1}}}}
\else
  \newcommand{\daniel}[1]{}
  \newcommand{\pablo}[1]{}
  \newcommand{\fix}[1]{}
\fi

\definecolor{cvprblue}{rgb}{0.21,0.49,0.74}
\usepackage[pagebackref,breaklinks,colorlinks,allcolors=cvprblue]{hyperref}
\usepackage[capitalize]{cleveref}

\def\paperID{44} % *** Enter the Paper ID here
\def\confName{CVPR}
\def\confYear{2025}

\title{Towards Continuous Home Cage Monitoring: An Evaluation of Tracking and Identification Strategies for Laboratory Mice}

\author{Juan Pablo Oberhauser\\
TLR Ventures\\
Redwood City, CA \\
{\tt\small pablo.oberhauser@murine.org}
\and
Daniel Grzenda\\
University of Chicago\\
Chicago, IL\\
{\tt\small grzenda@uchicago.edu}
}

\begin{document}
\maketitle

\begin{abstract}
    Continuous, automated monitoring of laboratory mice enables more accurate data collection and improves animal welfare through real-time insights. Researchers can achieve a more dynamic and clinically relevant characterization of disease progression and therapeutic effects by integrating behavioral and physiological monitoring in the home cage. However, providing individual mouse metrics is difficult because of their housing density, similar appearances, high mobility, and frequent interactions. To address these challenges, we develop a real-time identification (ID) algorithm that accurately assigns ID predictions to mice wearing custom ear tags in digital home cages monitored by cameras. Our pipeline consists of three parts: (1) a custom multiple object tracker (MouseTracks) that combines appearance and motion cues from mice; (2) a transformer-based ID classifier (Mouseformer); and (3) a tracklet associator linear program to assign final ID predictions to tracklets (MouseMap). Our models assign an animal ID based on custom ear tags at 30 frames per second with 24/7 cage coverage. We show that our custom tracking and ID pipeline improves tracking efficiency and lowers ID switches across mouse strains and various environmental factors compared to current mouse tracking methods.
\end{abstract}
    
\section{Introduction}
\label{sec:intro}

Advancements in sensor technologies and computational capabilities present new opportunities to enhance animal studies through continuous behavioral and physiological monitoring in the home cage. This approach enables more accurate data collection, improves animal welfare, and provides a dynamic, clinically relevant view of disease progression and therapeutic effects.

However, existing methods such as RFID tracking~\citep{catarinucci2013smart,fong2023pymousetracks}, photo beam interruption~\citep{spink2001ethovision}, and camera-based systems~\citep{giancardo2013automatic}  have limitations—failing to capture complex behaviors, relying on visible light, or lacking continuous monitoring. Additionally, current tools struggle to isolate individual behavior in group-housed mice and often are limited to monitoring during the light cycle, overlooking key nocturnal behaviors of mice.

The Envision\textsuperscript{TM} platform~\citep{robertson2024systemspaper} was developed using local hardware for data acquisition in cages (DAX), computer vision, cloud-based storage and analysis, and a readily accessible user interface to provide real-time insights into animal studies.

Attribution of behavior to individual mice in home cage monitoring relies on a robust, fast, and accurate identification system. We present an approach to use detection algorithms to track and identify mice wearing custom ear tags. This identification algorithm identifies three mice, wearing three different custom ear tags, within a custom home cage environment at 30 frames per second for 24/7 monitoring. 

Tracking and identifying mice in their home environment requires an accurate pipeline that can handle heavy occlusion, a variety of bedding types, enrichment, supportive care (i.e., HydroGel®), as well as food and water equipment. We also present an ID algorithm that combines tracking-by-detection with ID classification based on ear tags and demonstrates strong performance across environmental conditions. 

Our results show that combining a custom tracking-by-detection algorithm with a robust ID model achieves state-of-the-art performance on a dataset of multi-housed mice in home cages across a variety of environments. 

\begin{figure}[t]
    \centering
    \includegraphics[width=0.8\linewidth]{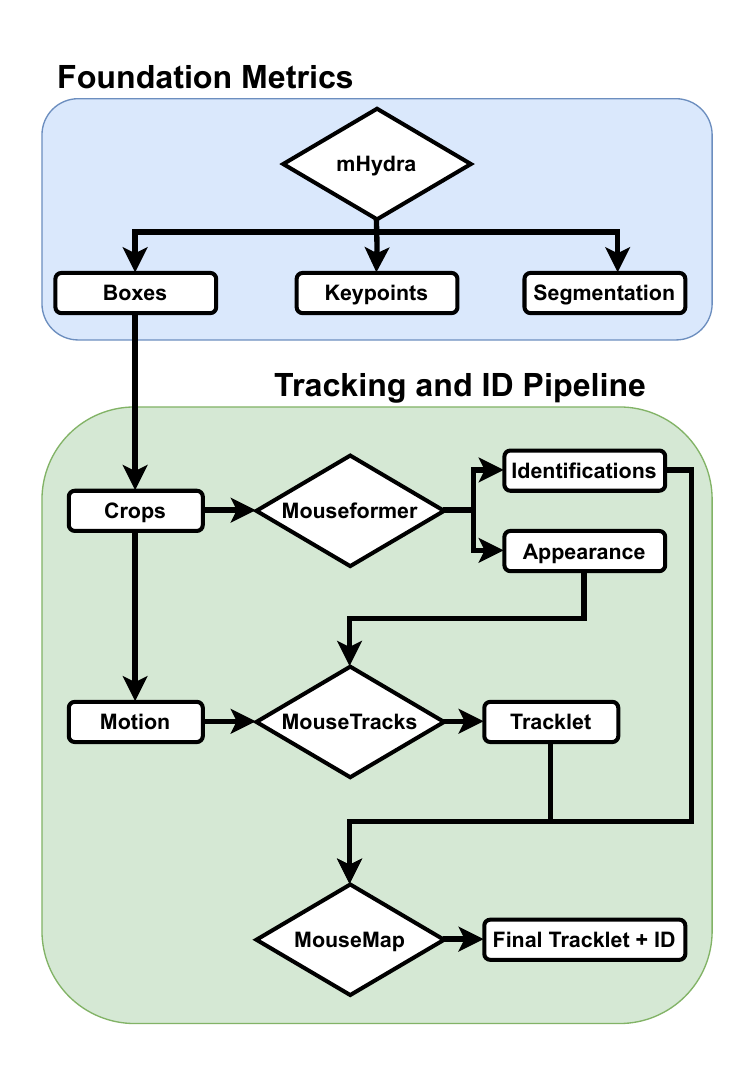}
    \caption{
        An overview of the tracking and identification pipeline developed for the Envision system. Foundational metrics are calculated before the Tracking and ID pipeline runs. 
    }
    \label{fig:diagram}
\end{figure}

\section{Related Work}
\label{sec:related}

\textbf{Tracking} Using appearance cues and motion cues to track by detection 
is a well-known paradigm in multi-object tracking. Most trackers in online tracking by detection paradigm rely on either motion cues~\citep{zhang2022bytetrack, cao2023observation, zhou2020tracking, zhang2021fairmot}, appearance cues~\citep{ veeramani2018deepsort, seidenschwarz2023simple, wang2020towards, cheng2016person, quan2019auto}, or a combination of both. Some projects aim at handling matching cascades~\citep{cao2023observation}, modeling motion explicitly with priors~\citep{liao2024enhanced,bergmann2019tracking,wu2021track}, 
or end-to-end learning~\citep{meinhardt2022trackformer}. There is also previous work on applying these tracking paradigms specifically to animal tracking~\citep{lauer2022multi, pereira2022sleap, kaul2024damm, mathis2018deeplabcut}.

\medskip
\noindent \textbf{Identification} 
Mice classification from video is a well-studied problem. Methods exist to identify individuals by shaving patterns in fur~\citep{cadillac2006animal}, tail-tattoos~\citep{deacon2006housing}, ear punching~\citep{roughan2019welfare}, and RFID implants~\citep{catarinucci2013smart}. There are also methods that rely solely on image data from the cages. Re-ID models identify animals by directly classifying the animal's identity in the crop using embeddings from a shared backbone \citep{pereira2022sleap, lauer2022multi}. Although Re-ID models are faster to run than a dedicated ID model, we saw much higher universal ID accuracy with a dedicated model and physical ear tags.

\section{Methods}
\label{sec:methods}

Below we organize our main contributions for tracking and identification of mice into three sections: (1) our tracking algorithm called MouseTracks, which uses detections and ID information; (2) our custom ID algorithm called Mouseformer, which uses custom ear tags and mouse crops to generate ID information; and (3), we characterize our constraint solver called Mousemap, which generates our final tracklets and ID assignments.

%The bedding provides significant challenges since mice and mouse parts can easily hide under the bedding.  

\subsection{Tracklet Generation}

\subsubsection{MouseTracks}

To link individual detections into tracklets, we use a custom tracker for mice named MouseTracker. MouseTracker aims to link individual detections into a set of temporally coherent detections, called a tracklet. Our custom tracker is heavily inspired by StrongSort~\citep{du2023strongsort} and DeepSort \citep{wojke2017simple} to combine appearance and motion cues into an online multi-object tracker.

To combine individual detections into trackets over time, we leverage motion and appearance cues in a video to link individual detections. This follows from the assumptions that detection boxes close to each other temporally belong to the same identity and that detection crops with similar visual appearances from one frame to another tend to belong to the same identity.  

\subsubsection{Motion Cues}
\label{subsubsec3}

To leverage motion cues, we use a sequence of detection boxes made by our mHydra~\citep{robertson2024systemspaper} model as inputs. The tracking paradigm is a standard SORT-style~\citep{wojke2017simple} tracker that uses a Kalman Filter to estimate the location of a box in a future step $t+1$ given momentum, velocity, and a history of locations from timestep $t$. The motion part of the cost matrix is derived from the IoU of a detection box to all other previous tracklets.

The state vector for the Kalman filter is a traditional bounding box tracker. As introduced by StrongSort~\citep{du2023strongsort}, we use the detection confidence scores to directly influence the update step. If the detection has low confidence, the updated state will be more influenced by its state estimate from the previous step. Conversely, if the detection is high confidence, the revised estimate will rely more on the state's measurement at the current step. 

\begin{figure}[t]
    \centering
    \includegraphics[width=0.8\linewidth]{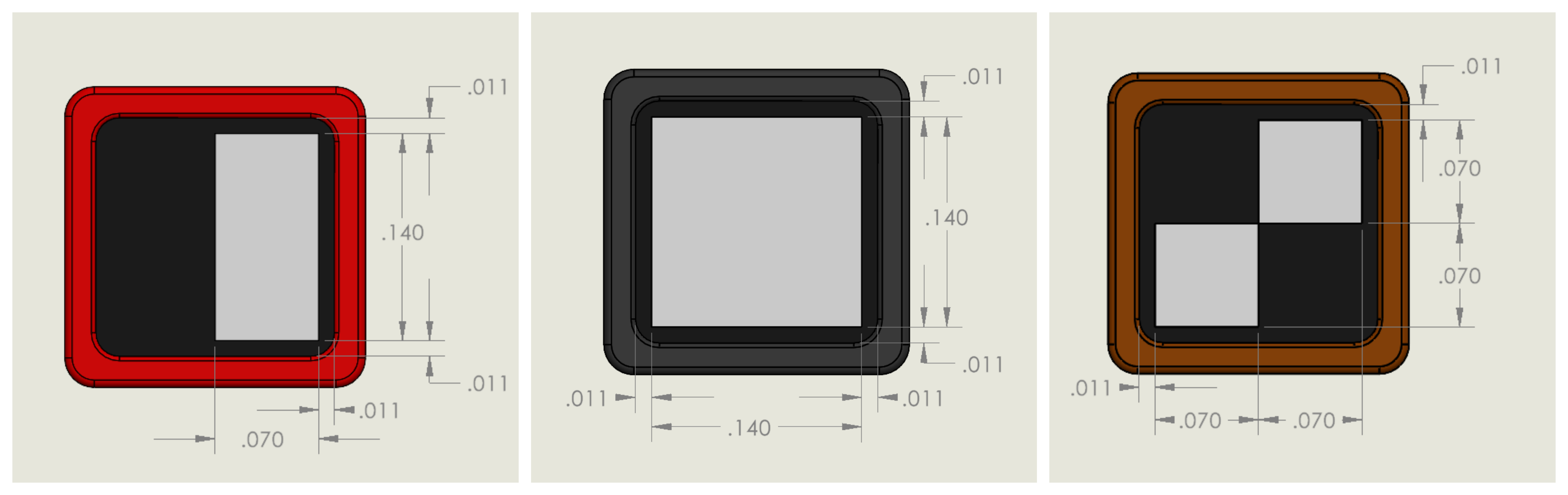}
    \caption{
        Custom ear tags for mouse identification. Patterns used on the custom RapID® ear tags for animal identification. Each ear tag has a pattern made of four squares that can be black (empty) or white (filled). Each ear tag is named for the background color and the pattern type. In the case of triply-housed mice, we use the following patterns, from left to right: red barred, black all-filled, and brown checkered.
    }
    \label{fig:eartags}
\end{figure}

There are, of course, other approaches we could take to leverage motion cues, like variations of IoU, including gIoU~\citep{rezatofighi2019generalized} or sIoU~\citep{gevorgyan2022siou}. The mHydra model also performs instance segmentation and pose estimation, all of which are outputs we could leverage to inform the motion cues in the tracker. We found pose estimation to be a noisier foundation metric for motion cues due to low confidence from occlusions and environmental variability across mouse cages. We leave the investigation of improving these methods for keypoints and instance segmentation to future work.

% The other tracker approach we experimented with was to replace using bounding boxes to calculate the motion cues with the detected pose vectors from the mHydra model. A series of direction vectors could be captured using combinations of keypoints. An example of a vector would be from the base of the tail keypoint to the tip of the nose keypoint. These vectors can be combined to hint at the pose and the direction that the mouse is moving in at any given point in time. Finally these vectors would be combined and tracked with a Kalman filter by using its start and end points as state estimators. This approach worked very well when the underlying pose detector is robust. Despite this promising direction we found pose estimation to be a noisier foundation metric for motion cues due to low confidence from occlusions and environmental variability across mouse cages. 

\subsection{Mouse Identification}

\begin{figure}
    \centering
    \begin{minipage}{0.48\linewidth}
      
      \begin{subfigure}{\linewidth}
        \includegraphics[width=\linewidth]{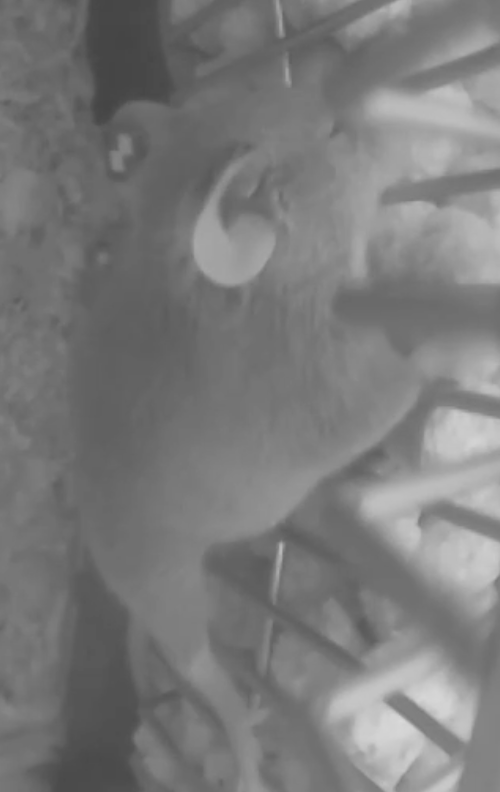}
          \caption{Brown checkered example}
          \label{fig:NiceImage1}
      \end{subfigure}

      \begin{subfigure}{\textwidth}
        \includegraphics[width=\textwidth]{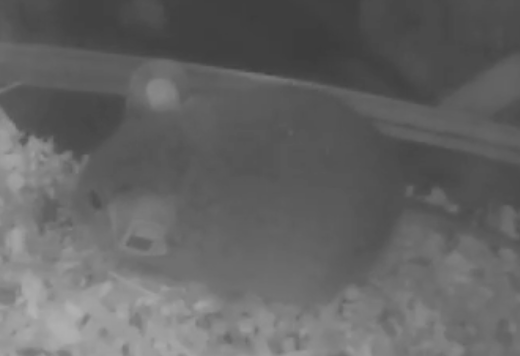}
          \caption{Red barred example}
          \label{fig:NiceImage2}
        \end{subfigure}
      
    \end{minipage}
    \begin{minipage}{0.48\linewidth}
    
      \begin{subfigure}{\textwidth}
        \includegraphics[width=\textwidth]{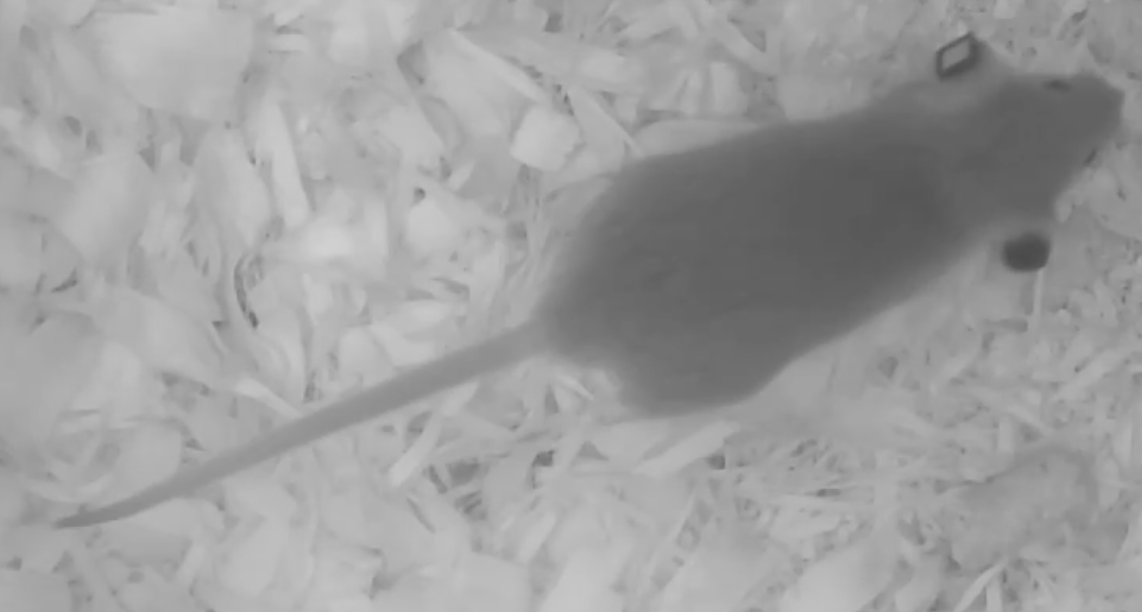}
          \caption{Black all filled example}
          \label{fig:NiceImage3}
      \end{subfigure}

      \begin{subfigure}{\textwidth}
        \includegraphics[width=\textwidth]{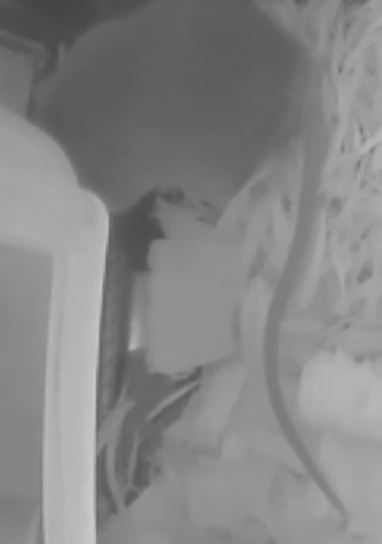}
          \caption{No read example}
          \label{fig:NiceImage3}
      \end{subfigure}
      \end{minipage}

\caption{
\label{fig:mousecrops}%
Examples of triple-housed mice in the Envison system using a DAX3 camera with custom RapID® barcoded ear tags. Crops are taken automatically from the bounding boxes outputs from the mHydra model.}
\end{figure}

To correctly assign an identity to a tracklet, tracklets are classified as one of the classes of ear tags (see Figure \ref{fig:eartags}) given a crop of a mouse (see Figure \ref{fig:mousecrops}).

There are two main challenges to assigning an ID to mouse crops. The first is that each mouse crop will be very similar, with the only difference being that a very small patch of the image contains the eartag (all the information needed to classify the image). The second challenge is that for 33\% of all objects detected, the ear tags are either not visible or have fallen off completely.

\subsubsection{Appearance Cues}
\label{subsubsec3}

For any set of three mice housed together, we expect the appearance of an animal at time $t$ is likely very similar to that of that animal at time $t-1$. Furthermore, we assume the appearance of animal A at time $t$ is likely more similar to that of Animal A at time $t-1$ than that of Animal B or C at time $t-1$. We expect the second assumption to be weaker due to the clonal nature of mice colonies. 

Therefore, we combine appearance and motion cues into a single cost matrix and use a simple Hungarian Algorithm to find optimal matches at every time step. Weighing the influence of these cues is done using a hyperparameter $lambda$ that can be tuned. The best weights for our use case are 0.9 for motion cues and 0.1 for appearance cues. 

To utilize appearance cues, we need to discriminate between animals that are very similar in appearance accurately. While a dedicated neural network could be used to generate numerical representations of appearance, we already have a model that can provide these appearance cues. By using the representation from our classifier Mouseformer, we reduce the runtime of our overall pipeline. The Mouseformer model produces both ear tag classification predictions and a set of vectors for each animal crop in the video that are fed into the tracking step. 

For each tracklet, we keep a single appearance vector that is the EMA of the last observations of that tracklet, first introduced in ~\citet{wang2020towards} and then used in \citet{du2023strongsort}.  The appearance cues from a new frame are normalized to the unit vector before being directly compared to these tracklet features using cosine similarity. Additional changes were made to optimize our model for mouse tracking. These included removing the GSI (Gaussian-smoothed interpolation) module, removing the ECC camera motion compensation, and replacing the AFLink linker with our final tracklet creator, MouseMap (described in \ref{sec6}).

\subsubsection{Ear Tag Classification}
\label{subsubsec2}

Fine-grained image classification aims to classify images from the same category into subcategories based on slight differences. In this case, the task is to classify images of mice in home cages into their ear tag class. This involves finding the most discriminative features associated with the ear tag pixels, likely subtle changes or hard-to-spot features, while ignoring macro environmental differences in the crop. For ID classification, a vision transformer is trained for fine-grained visual classification~\citep{diao2022metaformer, zheng2019looking}. We chose to use a CoAtNet backbone~\citep{dai2021coatnet}, which combines attention and convolutions for images. The CoAtNet is pre-trained on ImageNet-21k~\citep{ridnik2021imagenet}.

% There is a line of research that builds on the above to combine two inputs of data one being the image itself and the other being any cage metadata that can be given as text. We opt for a MetaFormer~\citep{diao2022metaformer} architecture to allow a model to jointly learn from cage annotations in the future. For now our architecture excludes text information making it
% To train the eartag classifier, we built a custom GUI that uses tracklets derived from our tracker and allows users to label ground truth eartrag class to train the Mouseformer. This GUI plays a video with tracklets overlaid and users can add grown truth. For example, if the video shows a clear tracklet with an MOT ID of 3, we can label that whole tracklet as "brown checkered" and the result is a set of mouse crops that can be used as training data. 

\subsection{Combining Tracklets and Identifications}\label{sec6}

The final step for our tracking and ID pipeline is to take in tracklets with corresponding ear tag class predictions (per frame) and assign a final, predicted ID to each tracklet. This is what we call MouseMap. At this step, we have a tracklet represented by $n$ detection boxes with $m$ unique classifier predictions and confidences from the five classes of ear tags. We use a constraint programming algorithm~\citep{martello1987linear} that uses a solver to provide the optimal assignment of tracklets to ID. We use the sum of the confidence of each ear tag classifier prediction in a tracklet as our objective function. 

The colloquial name for this kind of optimization problem is the hotel room problem, which maximizes profit given a set of requests for bookings on a finite set of rooms. In the same vein, we try to optimize the preference score over a finite set of identities (in this case, three). The $cpmpy$ \citep{guns2019increasing} library in Python makes it easy to model these constraint optimization problems with a solver, given an objective. We treat this problem as a utility maximization problem where the tracklets, defined by a start and end index in the video, all compete for allocation into one of the identities in the cage.

There are edge cases where the solver constraints are impossible to fulfill. For example, when there are more tracklet hypotheses than the number of mice in a cage. This happens since the multi-object tracker is unbound in the number of hypotheses it can create. When these cases arise, a final pre-solver step is triggered to give the three most probable tracklet hypotheses and merge shorter tracklet hypotheses. This merge step uses stitching to merge hypotheses that overlap in space but not in time (a break in tracking).

\section{Empirical Evaluation}
\label{sec:results}

\subsection{Datasets}

\textbf{Training} MouseFormer is the only module of this approach that requires training data, MouseTracks and MouseMap require no training data nor fine-tuning. MouseFormer is trained on a dataset of mouse crops and identities, given by the eartag that each animal is outfitted with. The data is used to train MouseFormer on an ID (classification) task. 

The source of the images for the MouseFormer classification set is gathered using the Envision platform, which records mice in home cages, 24/7 at 30 FPS. Our training dataset consists of 86,513 mouse crops. These crops are randomly sampled across various studies with different environmental conditions. Each crop is labeled with one of five eartag classes: brown-checkered (23\%), red-barred (21\%), black-all-filled (23\%), noread (17\%), and no-eartag (16\%). 

\begin{figure}[t]
    \centering
    \includegraphics[width=0.85\linewidth]{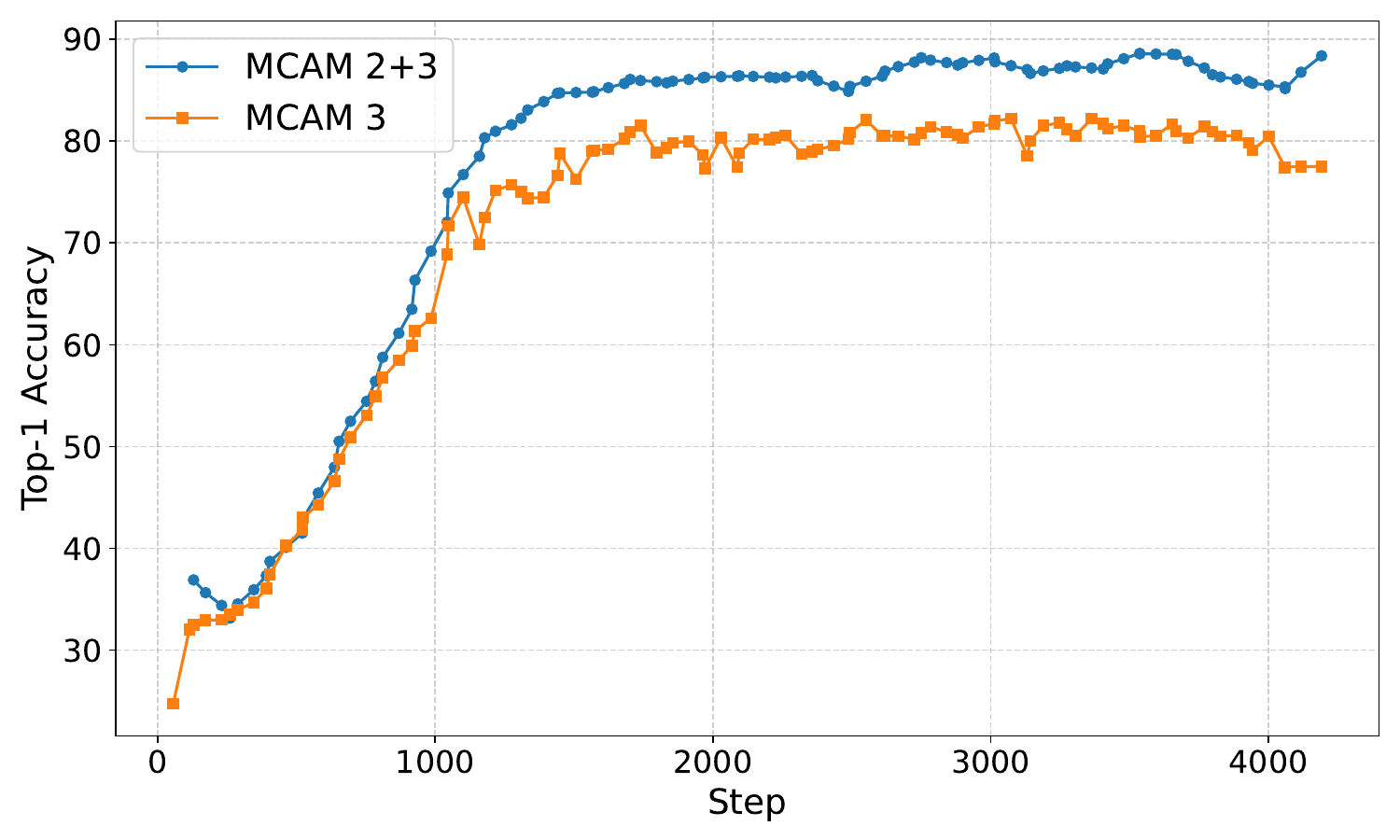}
    \caption{
        Performance of ID classification models across two different datasets. One dataset includes images from two different camera hardware (DAX2 and DAX3), and the other is of DAX3 data only. Both classifiers were evaluated on test data from DAX3. 
    }
    \label{fig:mcam}
\end{figure}

The dataset uses two camera hardware versions to train the Mouseformer: DAX2 and DAX3. When evaluating the Mouseformer solely on unseen DAX3 test data, we found that training with both DAX2 and DAX3 data in the training set performed best (see Figure~\ref{fig:mcam}). This highlights the importance of larger training sets for attention-based vision models~\citep{dosovitskiy2020image}, even if the training data is from a different distribution.

\medskip
\noindent 
\textbf{Validation} We have labeled 100 minutes of video with three mice in various environmental conditions. Data was gathered using DAX3 cages. We selected these 100 minutes to cover a wide range of environmental conditions, including contrasting mouse strains/coat colors (C3HJ/HeJ, C57BL/6J, and BALB/cJ), bedding (Alpha-dri®, Aspen Chips, Sani-chips), enrichment (wood blocks) nesting materials (4-8 grams brown shredded paper and 1-2 cotton squares), levels of occlusion, and lighting conditions (both light and dark cycle). Each mouse wears two custom RapID® barcoded ear tags (see Figure \ref{fig:eartags}). One tag was placed in each ear of each mouse, with one tag positioned anteriorly and the other positioned posteriorly. Our evaluation dataset consists of 100 minutes of data (179,980 frames with 524,663 objects) to be tracked and identified. We have both ID and tracking ground truth for each labeled minute. 

\subsection{Experiments}

Using these two datasets, we compare our entire ID and tracking pipeline to current state-of-the-art approaches for animal tracking. To do this, we compare our Mouseformer, MouseTracker, and MouseMap to both SLEAP~\citep{pereira2022sleap} and DeepLabCut~\citep{mathis2018deeplabcut}. We also evaluate different detectors' impact on these methods by influencing the tracklet and motion scores captured from each frame.

\begin{table}[t]
\begin{tabular}{lccc}\toprule

Detector + Tracker & IDF1 & MOTA & \# Switches \\\midrule
SA~\citep{ye2024superanimal} + DLC~\citep{mathis2018deeplabcut} & 72.33 & 77.21 & 24.5 \\
SA~\citep{ye2024superanimal} + SLEAP~\citep{pereira2022sleap} & 76.16 & 84.54 & 21.89 \\
mH~\citep{robertson2024systemspaper} + SLEAP~\citep{pereira2022sleap}& 67.59 & 91.03 & 11.94 \\
mH~\citep{robertson2024systemspaper} + Envision (ours) & \textbf{88.01} & \textbf{91.92} & \textbf{2.06} \\
\bottomrule

\end{tabular}

\caption{Detector and Tracker combined performance on 100 minutes of held-out dataset. Here, mH is mHydra detector, SA is the SuperAnimal detector, DLC is the DeeplabCut tracker, and SLEAP is the SLEAP tracker. We show that our tracker and detector combination produce longer and more stable tracklets while maintaining less ID switches per minute.}
\label{tab:results}
\end{table}

Our method using mHydra and the Envision tracker outperforms all other methods we tested (see Table \ref{tab:results}). We find a significant benefit to using our custom detector, as seen by comparing SLEAP's performance using the SuperAnimal detector to its performance using the mHydra detector. We also see a significant benefit in the number of ID switches from having a detected model for identification (Mouseformer) and using the appearance cues from those mouse crops to generate longer tracklets with fewer switches. Our method performs with an overall ID accuracy of 95.28\%.

\section{Conclusion}
\label{sec:conclusion}

We describe an animal identification and tracking pipeline that enables 24-hour monitoring of laboratory mice in home cages. We achieve state-of-the-art multi-object mice tracking and identification by incorporating video monitoring with custom physical ear tags. We developed this pipeline to deliver individual animal biomarkers in downstream scientific experiments. Future work on the pipeline will include addressing the challenges of scaling to multi-object tracking of five mice and removing the requirement for custom ear tags in favor of other mouse identification methods (e.g., tail tattoos or shaving patterns).

\section{Ethics}
The footage for both the training and validation sets come from the Envision\textsuperscript{TM}  by the Jackson Laboratory, an advanced digital in vivo monitoring system designed to assess mouse behavior in the home cage environment. All data was collected from animals under approved Institutional Animal Care and Use Committee (IACUC) protocols. 
\section{Acknowledgments}
This work was led and funded by The Jackson Laboratory. 

\clearpage

% \nocite{*}
{
    \small
    \bibliographystyle{ieeenat_fullname}
    \bibliography{main}
}


\end{document}